\documentclass[runningheads]{llncs}
\usepackage[T1]{fontenc}
\usepackage{graphicx}
\usepackage{multicol}
\usepackage{multirow}
\usepackage{booktabs}
\usepackage{algorithmic}
\usepackage{algorithm}
\usepackage{url}
\usepackage{amssymb}
\usepackage{amsmath}
\usepackage[misc]{ifsym}

\begin{document}

\title{Leveraging Queue Length and Attention Mechanisms for Enhanced Traffic Signal Control Optimization}

\titlerunning{Leveraging Queue Length and Attention Mechanisms for Enhanced TSC}
%
%

\toctitle{Leveraging Queue Length and Attention Mechanisms for Enhanced Traffic Signal Control Optimization}

\author{
Liang Zhang\orcidID{0000-0001-9683-7718} \and 
Shubin Xie\orcidID{0000-0001-6696-8418} \and 
Jianming Deng\thanks{Corresponding author}\Letter\orcidID{0000-0002-5901-2175}}

\authorrunning{L. Zhang et al.}
%
\tocauthor{Liang Zhang, Shubin Xie, Jianming Deng}

\institute{
State Key Laboratory of Herbage Improvement and Grassland Agro-ecosystems, College of Ecology, Lanzhou University, Lanzhou 730000, China \\
\email{\{liangzhang21, xieshb20, dengjm\}@lzu.edu.cn } 
}

\maketitle              
\begin{abstract}
Reinforcement learning (RL) techniques for traffic signal control (TSC) have gained increasing popularity in recent years.
However, most existing RL-based TSC methods tend to focus primarily on the RL model structure while neglecting the significance of proper traffic state representation. 
Furthermore, some RL-based methods heavily rely on expert-designed traffic signal phase competition.
In this paper, we present a novel approach to TSC that utilizes queue length as an efficient state representation.
We propose two new methods:
(1) Max Queue-Length (M-QL), an optimization-based traditional method designed based on the property of queue length;
and (2) AttentionLight, an RL model that employs the self-attention mechanism to capture the signal phase correlation without requiring human knowledge of  phase relationships.
Comprehensive experiments on multiple real-world datasets demonstrate the effectiveness of our approach: 
(1) the M-QL method outperforms the latest RL-based methods; 
(2) AttentionLight achieves a new state-of-the-art performance; 
and (3) our results highlight the significance of proper state representation, which is as crucial as neural network design in TSC methods.
Our findings have important implications for advancing the development of more effective and efficient TSC methods.
Our code is released on Github (\url{https://github.com/LiangZhang1996/AttentionLight}).

\end{abstract}
\section{Introduction}
With the growth of population and economy, the number of vehicles on the road has surged, leading to widespread traffic congestion. 
This congestion causes fuel waste, environmental pollution, and economic losses.
Enhancing transportation efficiency and alleviating traffic congestion has become crucial.
Signalized intersections are common bottlenecks in urban areas, and traffic signal control (TSC) is critical for effective traffic management.
Common TSC systems in modern cities include FixedTime~\cite{fixedtime}, GreenWave~\cite{greenwave}, SCOOT~\cite{scoot2}, and SCATS~\cite{scats2}.
%
%
%
These systems predominantly rely on expert-designed traffic signal plans, making them unsuitable for dynamic traffic and various intersections.

Reinforcement learning (RL)~\cite{rl}, a branch of machine learning, focuses on how intelligent agents should take action within an environment to maximize cumulative rewards. 
RL has attracted increasing attention for TSC, with researchers applying RL to address the TSC problem~\cite{drl,intellilight,LIT,cooperative,gcn,presslight,colight,mplight}.
Unlike traditional TSC methods, RL models can directly learn from the environment through trial and reward without requiring strict assumptions.
Furthermore, deep neural networks~\cite{DQN2015} powered RL models can learn to manage complex and dynamic traffic environments. 
RL-based TSC methods~\cite{presslight,mplight,colight} become a promising solution for adapting the dynamic traffic.
RL-based methods such as PressLight~\cite{presslight}, FRAP~\cite{frap}, MPLight~\cite{mplight}, and CoLight~\cite{colight} have emerged as promising solutions for adapting to dynamic traffic.


The performance in RL-based approaches can be influenced by the model framework, state representation, and reward function design.  
FRAP~\cite{frap} develops a specific network that constructs phase features and models phase competition correlations to obtain the score of each phase, yielding excellent performance for TSC.
%
CoLight~\cite{colight} uses graph attention network~\cite{gats} to facilitate intersection cooperation, achieving state-of-the-art performance.
LIT~\cite{LIT} leverages the network from IntelliLight~\cite{intellilight}  with a simple state scheme and reward function, significantly outperforming IntelliLight.
PressLight~\cite{presslight}  further optimizes the state and reward using pressure, considerably surpassing LIT.
MPLight~\cite{mplight} improves the state representation from PressLight and adopts a more efficient framework FRAP~\cite{frap}, significantly improving PressLight.

Various traffic state representations are employed, but the most effective state representation remains unknown.
State representations for RL-based TSC differ considerably compared to RL approaches in Atari games~\cite{atari}.
In the TSC field, state representation mainly varies in terms of the number of vehicles~\cite{intellilight,LIT,metalight,presslight,frap,metalight,colight,hilight}, vehicle image~\cite{drl,intellilight}, traffic movement pressure~\cite{mplight}, queue length~\cite{gcn,intellilight}, average velocity~\cite{gcn},  current phase~\cite{intellilight,LIT,metalight,presslight,frap,metalight,colight,hilight},  and next phase~\cite{intellilight,hilight}; reward representation varies in:  queue length~\cite{frap,presslight,colight,hilight}, pressure\cite{mplight,presslight}, total wait time~\cite{intellilight,gcn,drl,hilight}, and delay~\cite{intellilight,drl,hilight}.
Some methods such as LIT~\cite{LIT} and PressLight~\cite{presslight}, employ a simple state and reward and outperform IntelliLight~\cite{intellilight}, even with the same neural network.
Although traffic state representation plays an essential role in RL models, most research focuses on developing new network structures to improve TSC performance. 
Consequently, state design for TSC merits further consideration.

Recent studies,  such as MPLight~\cite{mplight} and MetaLight~\cite{metalight}, have adopted FRAP as their base model.
However, FRAP necessitates manually designed phase correlations, such as competing, partial competing, and no competing relationships in a standard four-way and eight-phase (Figure~\ref{fig:inter}) intersection~\cite{frap}. 
While the analysis of phase and traffic movements can aid in determining phase correlations, this approach may be impractical for more complex intersections, such as five-way intersections.
%

To tackle the aforementioned challenges, this article  presents the following key contributions:
(1) we propose an optimization-based TSC method called Max Queue-Length (M-QL), and (2) develop a novel RL model, AttentionLight, which leverages self-attention to learn phase correlations without requiring human knowledge of phase relationships.
Extensive numerical experiments demonstrate that our proposed methods outperform previous state-of-the-art approaches, with AttentionLight achieving the best performance.
Additionally, our experiments highlight the significance of state representation alongside neural network design for RL.


\section{Related Work}

\subsection{Traditional Methods}
Traditional methods for  traffic signal control (TSC) can be broadly categorized into four types: fixed-time control~\cite{fixedtime}, actuated control~\cite{sotl2004,sotl2013}, adaptive control ~\cite{scoot2,scats2}, and optimization-based control~\cite{mp2013,mp2015,mp2018}.
Fixed-time control~\cite{fixedtime} utilizes pre-timed cycle length, fixed cycle-based phase sequence, and phase split, assuming uniform traffic flow during specific periods.
%
Actuated control~\cite{sotl2013} decides whether to maintain or change the current phase based on the pre-defined rules and real-time traffic data, such as setting a green signal for a phase if the number of approaching vehicles exceeds a threshold.
Self-organizing traffic lights (SOTL)~\cite{sotl2013} is one typical actuated control method.
Adaptive control ~\cite{scoot2,scats2} selects an optimal traffic plan for the current situation from a set of traffic plans based on traffic volume data from loop sensors. Each  plan includes cycle length, phase split, and offsets. 
SCOOT~\cite{scoot2} and SCATS~\cite{scats2} are widely used adaptive control methods in modern cities.
%
Optimization-based control~\cite{mp2013,mp2015,mp2018} formulates TSC as an optimization problem under a specific traffic flow model, using  observed traffic data to make decisions. 
Max Pressure~\cite{mp2013} is a typical optimization-based control method that often requires turn ratio (the proportion of turning vehicles at an intersection).

\subsection{RL-Based Methods}

Reinforcement learning (RL)-based methods have been employed to improve traffic signal control (TSC) performance, with several studies concentrating on optimizing state and reward design. 
A trend has emerged favoring simpler yet more efficient state representations and reward functions. 
For example, IntelliLight~\cite{intellilight} employed six state representations and six features to compute the reward function, resulting in moderate performance. 
In contrast, LIT~\cite{LIT} used the current phase and the number of vehicles as the state, and queue length as the reward, significantly outperforming IntelliLight. 
PressLight~\cite{presslight} further improved upon LIT and IntelliLight by incorporating 'pressure' into the state and reward function design.
MPLight~\cite{mplight} enhanced FRAP~\cite{frap} by adopting traffic movement pressure in the state and reward function design.

Other studies have focused on improving control performance by employing more powerful networks or RL techniques. 
For instance, FRAP~\cite{frap} developed a unique network structure to construct phase features and capture phase competition relations, resulting in invariance to symmetrical cases such as flipping and rotation in traffic flow. 
GCN~\cite{gcn} utilized graph convolution networks~\cite{gcn2} with queue length and average velocity as the state and total wait time as the reward. 
CoLight~\cite{colight} introduced graph attention network~\cite{gats} to facilitate intersection cooperation, using the number of vehicles and current phase as the state and queue length as the reward. 
HiLight~\cite{hilight} incorporated the concept of hierarchical RL~\cite{hierarchical}, using the current phase, next phase, and the number of vehicles as state, while employing queue length, delay, and waiting time as the reward.

Although numerous studies strive to develop complex network structures for TSC, few focus on appropriate traffic state representation design. 
LIT~\cite{LIT} demonstrated that queue length serves as a more effective reward function than delay, and the number of vehicles surpasses waiting time and traffic image as the state representation. 
PressLight~\cite{presslight} discovered that pressure outperforms queue length as the reward function. 
MPLight~\cite{mplight} integrated pressure into state design, resulting in improvements to the model. 
These findings indicate that further research on state representation is required to enhance TSC methods. 
Moreover, the exploration of novel network structures within RL techniques for TSC should be considered.


\section{Preliminary}

\begin{figure}[htb]
\centering
  \includegraphics[width=0.76\linewidth]{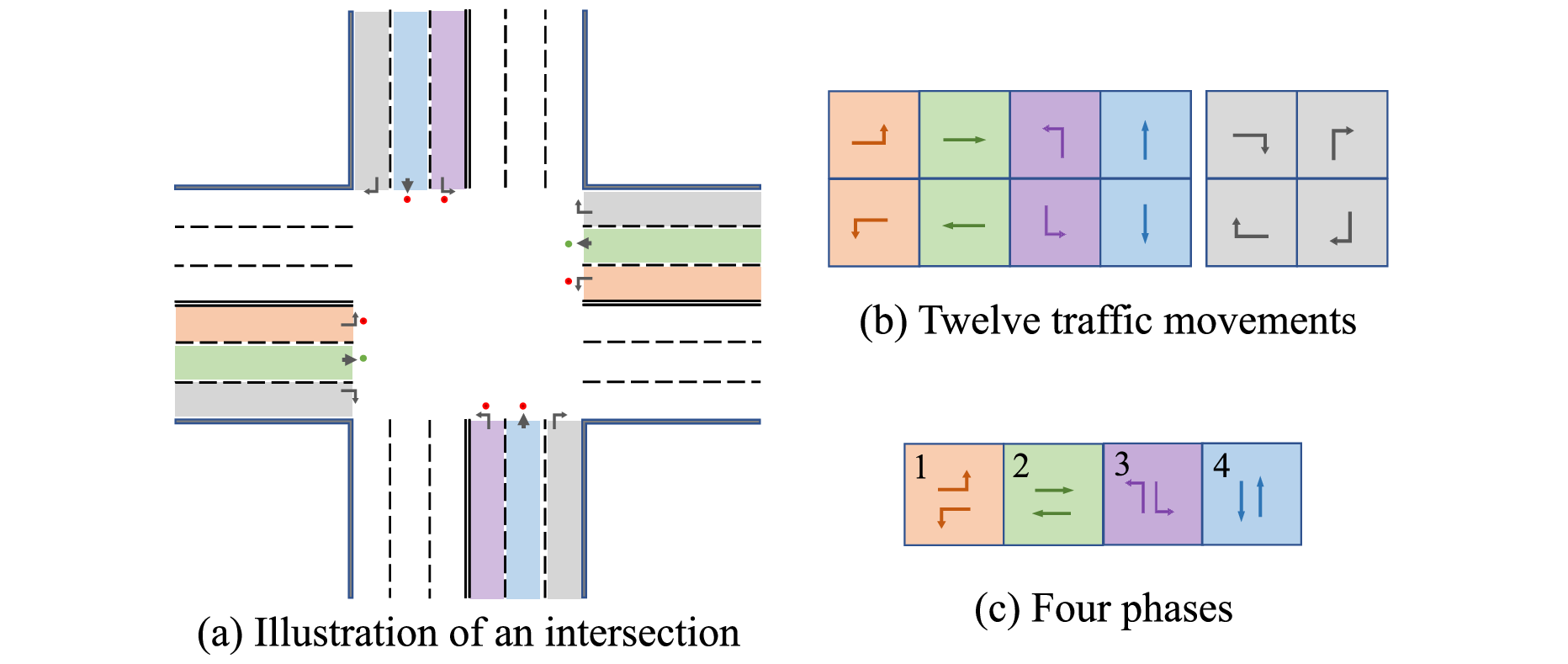}
  \caption{The illustration of an intersection. In case (a), phase \#2 is activated.}
  \label{fig:inter}
\end{figure}

In this study, our primary focus is on conventional and representative 4-way, 12-lane, 4-phase intersections (Figure~\ref{fig:inter}).
In this section, we provide a comprehensive summary of the definitions that are integral to TSC methods.

\textbf{Definition 1} (Traffic network). 
The traffic network is described as a directed graph where each node represents an intersection, and each edge represents a road. 
One road consists of several lanes with vehicles running on it.
%
%
We denote the set of incoming lanes and outgoing lanes of intersection $i$ as $\mathcal{L}_{i}^{in}$ and $\mathcal{L}_{i}^{out}$ respectively.
The lanes are denoted with $l,m,k$. As is shown in Figure~\ref{fig:inter} (a), there are one intersection, four incoming roads, four outgoing roads, twelve incoming lanes, and twelve outgoing lanes. 

\textbf{Definition 2} (Traffic movement). A traffic movement is defined as the traffic traveling across an intersection towards a specific direction, i.e., left turn, go straight, and right turn. 
According to the traffic rules in some countries, vehicles that turn right can pass regardless of the signal but must yield at red lights. As shown in Figure~\ref{fig:inter} (b), there are twelve traffic movements.

\textbf{Definition 3}(Signal phase). Each signal phase is a set of permitted traffic movements, denoted by $d$, and $\mathcal{D}_i$ denotes the set of all the phases at intersection $i$. 
As shown in Figure~\ref{fig:inter}, twelve traffic movements can be organized into four-phase (c), and phase \#2 is activated in case (a).


\textbf{Definition 4} (Phase queue length). The queue length of each phase is the sum queue length of the incoming lanes of that phase, denoted by 
\begin{equation}
	q(p) = \sum q(l), l \in p
\end{equation}
in which $q(l)$ is the queue length of lane $l$.

\textbf{Definition 5} (Intersection queue length). The queue length of each intersection is defined as the total queue length of the incoming lanes of the intersection, denoted by 
\begin{equation}
	Q_i = \sum q(l),  l \in \mathcal{L}_{i}^{in}
\end{equation}
in which $q(l)$ represents the queue length of lane $l$.

\textbf{Definition 6} (Action duration). The action duration of our TSC models is denoted by $t_{duration}$.
It can also represent the minimum duration of each phase.

\textbf{Problem} (Multi-intersection TSC). We consider multi-intersection TSC, in which each intersection is controlled by one RL agent.
Every $t_{duration}$, agent $i$ views the environment as its observation $o_i^t$, takes the action $a_i^t$ to control the signal of intersection $i$, and obtains reward $r_i^t$. 
Each agent can learn a control policy by interacting with the environment.
The goal of all the agents is to learn an optimal policy (i.e. which phase to actuate) to maximize their cumulative reward, denoted as:
\begin{equation}
\sum_{t=1}^{T}\sum_{i=1}^{n} r_i^t
\end{equation}
where $n$ is the number of RL agents and $T$ is the timestep.
The well-trained RL agents will be evaluated with multiple real-world datasets.
%
%


\section{Method}
In this section, we first propose a  TSC method, Max-QueueLength(M-QL), based on the property of queue length.
Next, we present a novel RL model called AttentionLight, which employs multi-head self-attention~\cite{attention} to model phase correlation and utilizes queue length both as the state and reward.

%

\subsection{Introduce Queue Length for TSC Methods}

\subsubsection{Property of Queue Length}
In TSC, vehicles within the traffic network can be in one of two states: moving or queuing.
Queuing vehicles can lead to congestion and are essential for representing the traffic condition.
%
The phase signal can directly change the state of the queuing vehicles in the traffic network.
%
A deterministic change transpires when each phase is activated, causing the queue length of that phase to decrease to zero.
%
In contrast, subsequent changes, such as the number of vehicles, vehicle position, and vehicle speed, exhibit greater uncertainty compared to queue length.
%


RL agents learn the state-action values from the environment through trial and reward. 
The feedback on actions significantly influences the learning effect.
Suppose the state representation omits critical contents of traffic movement. 
In such cases, agents may become confused about the state and fail to learn an appropriate policy for TSC.
For example, consider a scenario where  one case has only queuing vehicles and another has only moving vehicles, and the state representation is based solely on the number of vehicles.
%
Under the same state representation, there could be different optimal policies, which may confuse the RL agents.
Furthermore, the state space of queue length is larger than that of the number of vehicles.
As a result, the queue length is considered an effective state representation.
Using queue length as both the reward and the state can support reward optimization.
Consequently, we employ queue length as a traffic state representation and reward function.

While previous studies have incorporated queue length in state and reward design, our work is the first to use it as both.
IntelliLight~\cite{intellilight}, GCN~\cite{gcn}, and Tan et al.~\cite{cooperative} employ more complex state representations.
In contrast, our approach uniquely emphasizes simplicity and efficiency by focusing solely on queue length.

\subsubsection{Max-QueueLength Control}

Based on Max Pressure (MP)~\cite{mp2013} and the property of queue length, we introduce a new TSC method called Max Queue-Length (M-QL), which directly optimizes intersection queue length.
%
The M-QL control chooses the phase that has the maximum queue length in a greedy manner.
At intersection $i$, the queue length of each phase is calculated using equation (1). 
During each action duration, M-QL activates the phase that has the maximum queue length, denoted by
\begin{equation}
\hat{p} = \operatorname{arg max}\left(q(p) p\in \mathcal{P}_i\right)
\end{equation}
where $q(p)$ is calculated according to equation (1), and $\mathcal{P}_i$ denotes the phases.
Our approach is straightforward and efficient, and we believe it has the potential to improve traffic conditions significantly.


\subsubsection{M-QL can stabilize the network} 
We present proof of the stability of M-QL control for TSC.

\textbf{Definition 7} (Queue length process stability)~\cite{mp2013}. The queue length process $Q(t)=\{q(l)\}$ is stable in the mean (and $u^{*}$ is a stabilizing control policy) if for some $M<\infty$:
\begin{equation}
    \frac{1}{T} \sum_{t=1}^{T}\sum_{l} \operatorname{E}[q(l)] < M, \forall T.
\end{equation}
where $\operatorname{E}$ denotes expectation.
Stability in the mean implies that the chain is positively recurrent and has a unique steady-state probability distribution.

\begin{theorem} The M-QL control $u^{*}$ is stabilizing whenever the average demand is admissible\footnote{An admissible demand means the traffic demand can be accommodated by traffic signal control policies, not including situations like long-lasting over-saturated traffic that requires perimeter control to stop traffic getting in the system.}.
\end{theorem}
\begin{proof}  
We use $x(l)$ and $x_{max}(l)$ to denote the number of vehicles and the maximum permissible vehicle number on lane $l$, respectively.
Based on the property of queue length, we have $q(l) \leqslant x(l)$.
Moreover, as the average demand is admissible, $x(l) \leqslant x_{max}(l)$.
For a rough estimation, we obtain $M \leqslant \frac{1}{T} \sum_{t=1}^{T}\sum_{l} \operatorname{E}[x_{max}(l)]$.

For the traffic conditions in this study, such as shown in Figure~\ref{fig:inter} (a), there are four phases and twelve lanes.
Vehicles that turn right can pass regardless of the signal.
Additionally, M-QL always actuates one phase, and there are no queuing vehicles on the lane of that corresponding phase.
Hence, there are no queuing vehicles on half of the lanes. 
For a more precise estimation, we can get $M \leqslant \frac{1}{T} \sum_{t=1}^{2T}\sum_{l} \operatorname{E}[x_{max}(l)]$.

Therefore, based on the stability criterion defined in Equation (5), we prove that the M-QL control policy is stabilizing, and the queue length process is stable in the mean whenever the average demand is admissible.
\end{proof}

\subsubsection{Comparison of M-QL and MP}
The MP control selects the phase with the maximum pressure, which is the difference in queue length between upstream and downstream, indicating the balance of the queue length. 
Similarly, the M-QL control opts for the phase with the maximum queue length in a greedy manner.
%
These two methods are identical for single intersection control, where the outgoing lanes are infinite, and the calculated pressure equals the queue length. 


However, MP also considers  the neighboring influences and stabilizes the queue length by ensuring that vehicles are not stopped by upstream queuing vehicles.
Consequently, a phase with higher pressure would result in a larger queue length.
MP is effective for short traffic road lengths, where the influence of adjacent intersections is felt rapidly.
Nevertheless, for long road lengths, pressure may not be as effective since the impact can be several $t_{duration}$ (action duration) away.
%
For example, for a road length of 300m, with a $t_{duration}$ of 15 seconds and a vehicle maximum velocity of $10m/s$, it would take at least 30 seconds to reach the neighbor, rendering the neighbor condition ineffective.
In contrast, M-QL might perform better on longer traffic roads, as it directly optimizes queue length. Experiments will be conducted later to verify this assumption.

\subsection{AttentionLight Agent Design}

Prior to delving into the network architecture of AttenditonLight, it is essential to elucidate  the state, action, and reward for each RL agent.
Considering queue length's role in M-QL, we suggest incorporating it in both state and reward design, expected to boost our model's performance and efficiency.
\begin{itemize}
    \item \textbf{State}. The current phase and queue length are used as the state representation (agent observations). The state at time $t$ is denoted as $s_t$. 
    \item \textbf{Action}. At the time $t$, each agent chooses a phase $\hat{p}$ according to the observations, and the traffic signal will be changed to $\hat{p}$. The action thus influences traffic flow by modifying the traffic signal. 
    \item \textbf{Reward}. The negative intersection queue length is used as the reward. The reward for the agent that is controlling intersection $i$ is denoted by
\begin{equation}
	r_i =-\sum q(l), l \in \mathcal{L}_i^{in}	
\end{equation}
in which $q(l)$ is the queue length at lane $l$. 
By maximizing the reward, the agent is trying to maximize the throughput in the system. 
In this study, we update our agent based on the average reward ($r_i$) over the action duration, taking into account the reward delay.

\end{itemize}

\paragraph{Advanced RL framework} The DQN~\cite{DQN2015} is used as the function approximator to estimate the Q-value function, and the RL agents are updated by the Bellman Equation.
We also employ a decentralized RL paradigm, including ApeX-DQN~\cite{apex}, for scalability.
This approach shares parameters and replay memory among all agents, enabling intersections to learn from each other's experiences, a technique proven to enhance model performance~\cite{mplight}.

\subsection{Network Design of AttentionLight}

Though some RL-related methods~\cite{frap,mplight,metalight} using FRAP~\cite{frap} have achieved impressive performance, they rely on human-designed phase correlations.
To overcome this limitation, we propose AttentionLight, an RL model based on FRAP that utilizes self-attention~\cite{attention} to automatically model phase correlations.

The core idea of AttentionLight is to apply self-attention~\cite{attention} to learn phase correlations and predict the Q-value of each phase through the phase feature constructed by the self-attention mechanism.
This approach enables the Q-value of each phase to fully consider its correlation with others.
We divide the prediction of Q-values (i.e. the score of each phase) into three stages: phase feature construction, phase correlation learning with multi-head self-attention, and Q-value prediction.

\textbf{Phase Feature Construction}
AttentionLight utilizes queue length and current phase as inputs.
Initially, AttentionLight embeds the feature of each lane from $l$-dimensional into a $m$-dimensional latent space via a layer of multi-layer perceptron:
\begin{equation}
    h_1 = \operatorname{Embed}(o) = \sigma(oW_e+b_e)
\end{equation}
where $o\in\mathbb{R}^{l}$ is the observation at time $t$, $W_e\in\mathbb{R}^{l\times m}$ and $b_e\in\mathbb{R}^{m}$ are weight matrix and bias vector to learn, $\sigma$ is the sigmoid function.
Subsequently,  the feature of each phase is constructed through feature fusion of the participating lanes:
\begin{equation}
    h_2 = \operatorname{Fusion}(h_1)
\end{equation}
in this case, the fusion function is a direct addition.

\textbf{Phase Correlation Learning} In this stage, our model takes the phase feature as input and uses multi-head self-attention(MHA)~\cite{attention} to learn the phase correlation: 
\begin{equation}
    h_3 = \operatorname{MHA}(h_2)
\end{equation}
we find that the head number does not have a significant influence on the model performance, and we finally adopt four attention heads as default.

\textbf{Q-Value Prediction} In this stage, our model takes the correlated phase feature as input to get the Q-value for each phase:
\begin{equation}
    \widetilde{q} = \operatorname{Embed}(h_3) =h_3 W_p + b_p
\end{equation}
where $W_p\in\mathbb{R}^{c\times 1}$ and $b_p\in\mathbb{R}^{1}$ are parameters to be learned, $p$ denotes the number of phases (action space), $\widetilde{q}$ refers to the predicted q-values.
The agent selects the phase that has the maximum Q-value.

\subsubsection{AttentionLight and FRAP}

AttentionLight is a novel RL model that uses self-attention to automatically model phase correlation for TSC.
Unlike FRAP~\cite{frap}, which requires human-designed phase correlation, AttentionLight does not rely on human knowledge of the complex competing relationships between phases. 
For instance, in a typical 4-way and 8-phase intersection,  phases can have competing, partial competing, or no competing relationships. 
Although the competing relationships of the 8 phases in such an intersection can be easily acquired through analysis of phase and traffic movements, this task becomes considerably more challenging for more complex intersections or phases. 

AttentionLight is better suited for real-world deployment than FRAP, as it significantly reduces the complexity of phase relation design. 
By learning the phase correlation through a neural network, our model enables scalability and eliminates the need for human intervention.

\section{Experiment}



\paragraph{Settings} 
We conduct comprehensive numerical experiments on CityFlow~\cite{cityflow}, where each green signal is followed by a five-second red time to prepare for the signal phase transition.
Within this simulator, vehicles navigate toward their respective destination following pre-defined routes, adhering strictly to traffic regulations.
Control methods are deployed to control the signals at each intersection to optimize the traffic flow. 
We evaluate our proposed methods using seven real-world traffic datasets~\cite{colight}\footnote{https://traffic-signal-control.github.io} sourced from JiNan, HangZhou, and New York.
These datasets have been extensively utilized by various methods, such as  CoLight~\cite{colight}, HiLight~\cite{hilight}, and PRGLight~\cite{prglight}.
The traffic networks under JiNan, HangZhou, and New York each exhibit unique topologies. 
In JiNan, the road network comprises 12 intersections ($3\times4$), each linking two 400-meter East-West and two 800-meter South-North road segments; in HangZhou, the road network consists of 16 intersections ($3\times4$), each connecting two 800-meter East-West and two 600-meter South-North road segments; in New York, the road network has 196 intersections ($28\times 7$), each connecting four 300-meter (two East-West and two South-North) road segments.
The average arrival rate (vehicles/second) of the seven datasets is 1.75, 1.21, 1.53, 0.83, 1.94, 2.97, and 4.41 respectively.
These traffic flow datasets not only vary in terms of arrival rate but also  in travel patterns, thereby demonstrating the diversity and validity of our experiments.

Drawing  upon prior research~\cite{frap,colight,hilight}, we select the average travel time as the evaluation metric and compare our methods with various traditional and RL approaches. 
To ensure a fair comparison,  we set the phase number as four(Figure~\ref{fig:inter} (c)), and the action duration as 15 seconds. 
All RL methods are trained using the same hyper-parameters, such as optimizer (Adam), learning rate (0.001), batch size (20), sample size (3000), memory size (12000), epochs number (100), discount factor $\gamma$ (0.8), etc.
In order to derive definitive results for all the RL methods, a total of 80 episodes are utilized.
Each episode, both in training and testing, executes a simulation lasting 60 minutes.
The mean value is then calculated based on the final ten testing episodes.
To bolster the reliability of our findings, we conducted three independent experiments and reported the average outcome.
This rigorous approach ensures the robustness and reproducibility of our results.

\paragraph{Compared Methods} The traditional methods include: \textbf{FixedTime}~\cite{fixedtime}, a policy that uses a fixed cycle length with a pre-defined phase split among all the phases; \textbf{Max Pressure (MP)}~\cite{mp2013}: a policy that selects the phase with the maximum pressure. 
The RL-based methods include: \textbf{FRAP}~\cite{frap},  which uses a modified network structure to capture phase competition relation between signal phases; \textbf{MPLight}~\cite{mplight}, which uses FRAP as the base model, incorporates pressure in the state and reward design, and has shown superior performance in city-level TSC; \textbf{PRGLight}~\cite{prglight}, which  employs a graph neural network to predict traffic state and adjusts the phase duration according to the current observed traffic state and predicted state; \textbf{CoLight}~\cite{colight}, which uses a graph attention network~\cite{gats} to facilitate intersection cooperation and has shown superior performance in large-scale TSC, making it a state-of-the-art method.

\subsection{Overall Performance}
Table~\ref{tab:all} reports the performance of all the methods under JiNan, HangZhou, and New York real-world datasets in terms of average travel time. 
Traditional TSC methods, such as MP~\cite{mp2013} and M-QL, continue to demonstrate competitive results.
%
Specifically, MP~\cite{mp2013} outperforms FRAP~\cite{frap}, MPLight~\cite{mplight}, PRGLight~\cite{prglight}, and CoLight~\cite{colight} under JiNan and HangZhou datasets.
Our proposed M-QL consistently outperforms all other previous methods under JiNan and HangZhou datasets, with an improvement of up to 4.21\% (averaging 2.47\%), while MP surpasses M-QL under New York datasets.
We hypothesize that the length of traffic roads influences the performance of TSC methods, with M-QL potentially excelling on longer roads and MP on shorter roads.
These results can validate this hypothesis.

\begin{table}[htb]
    \centering
    \caption{Overall performance. For average travel time, the smaller the better.}
    \label{tab:all}
    \begin{tabular}{lccccccc}
    \toprule
    \multirow{2}{*}{ Method } & \multicolumn{3}{c}{ JiNan } & \multicolumn{2}{c}{ HangZhou } & \multicolumn{2}{c}{New York} \\
    \cmidrule { 2 - 8 } & 1 & 2 & 3 & 1 & 2 & 1 & 2\\
    \midrule
    FixedTime & $429.27$ & $370.34$ & $384.89$ & $497.87$ & $408.31$ & $1507.12$ & $1733.30$ \\
    MP & $274.99$ & $\mathbf{246.41}$ & $\mathbf{244.63}$ & $\mathbf{289.54}$ & $349.85$ & $1179.55$ & $1536.17$ \\
    \midrule
    FRAP & $299.56$ & $268.57$ & $269.20$ & $308.73$ & $355.80$ & $1192.23$ & $1470.51$ \\
    MPLight & $297.68$ & $274.32$ & $268.00$ & $313.16$ & $355.35$ & $1321.40$ & $1642.05$ \\
    PRGLight & $ 291.27$& $ 257.52$& $ 261.74$& $ 301.06$& $ 369.98$& $ 1283.37$& $ 1472.73$\\
    CoLight & $\mathbf{271.17}$ & $251.22$ & $248.87$ & $300.07$ & $\mathbf{339.76}$ & $\mathbf{1065.64}$ & $\mathbf{1367.54}$ \\
    \midrule 
    M-QL & $\mathbf{268.87}$ & $\mathbf{240.02}$ & $\mathbf{238.51}$ & $\mathbf{284.32}$ & $\mathbf{325.44}$ & $1197.59$ & $1551.46$ \\
    AttentionLight & $\mathbf{254.82}$ & $\mathbf{239.68}$ & $\mathbf{236.62}$ & $\mathbf{283.64}$ & $\mathbf{316.38}$ & $\mathbf{1013.78}$ & $1401.32$ \\
    \bottomrule
    \end{tabular}
\end{table}

Additionally, our proposed AttentionLight achieves new state-of-the-art performance and outperforms all other previous methods over JiNan and HangZhou datasets, with an improvement of up to 6.88\% (averaging 4.34\%).
AttentionLight exclusively utilizes queue length information of a specific intersection, thus requiring less computation and  offering deployment advantages  over CoLight, MPLight, and FRAP. 
Finally, we reemphasize the importance of parameter sharing for RL-based models in TSC. 
MPLight~\cite{mplight} has demonstrated superior performance than FRAP and addressed the importance of parameter sharing.
When FRAP is trained and tested with parameter sharing in the same manner as MPLight, it slightly outperforms MPLight.

\subsection{Queue Length Effectiveness Analysis}

To further illustrate the effectiveness of queue length as a state representation, we incorporate queue length as both the state and reward for additional RL methods, including FRAP~\cite{frap} and CoLight~\cite{colight}, which are referred to as QL-FRAP and QL-CoLight, respectively.
%
Additionally,  we  introduce a simple DQN containing two multiple-layer perceptrons  as QL-DQN.

\begin{figure}[htb]
    \centering
    \includegraphics[width=0.8\linewidth]{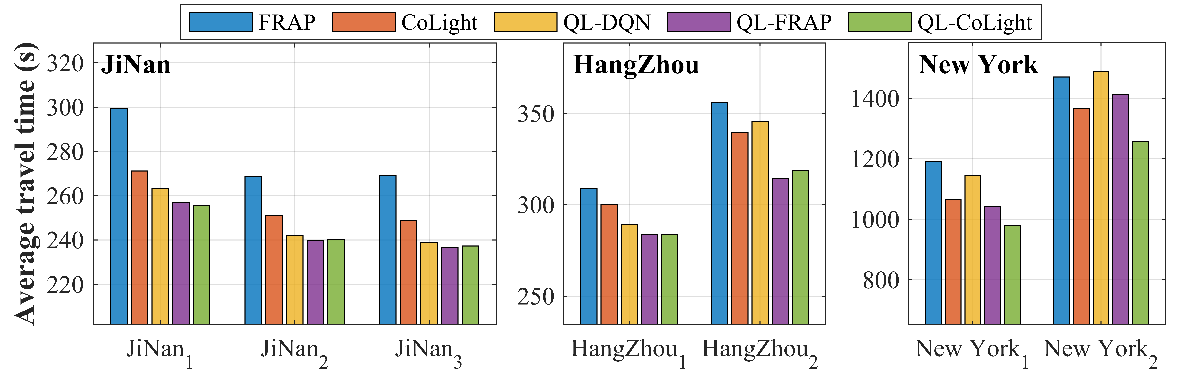}
    \caption{Model performance comparison.}
    \label{fig:efficient}
\end{figure}

Figure~\ref{fig:efficient} demonstrates the model performance of QL-DQN, QL-FRAP, and QL-CoLight.
The results show that QL-FRAP significantly outperforms FRAP, and QL-CoLight significantly outperforms CoLight.
These improvements highlight the importance of state representation for RL-based TSC.
Moreover, QL-DQN outperforms FRAP and CoLight, further emphasizing the critical role of state representation in RL.
Efficient state representation is also essential as the neural network structure for TSC.
QL-DQN employs a simple neural network structure, but efficient state representation.
In contrast, FRAP and CoLight use well-designed neural network structures but have less efficient state representation.
Furthermore, FRAP, CoLight, and QL-DQN use the same reward function.
When comparing the performance of QL-DQN with FRAP and CoLight, QL-DQN consistently performs better under JiNan and HangZhou datasets.
These experimental results suggest that queue length serves as an efficient state representation.


\subsection{Reward Function Investigation}

Previous studies, such as PressLight~\cite{presslight} and MPLight~\cite{mplight}, have demonstrated the superior performance of RL approaches under pressure compared to queue length in the context of reward functions.
This study revisits these findings, focusing on the impact of reward settings when queue length is used as the state representation.
We utilize two base models for our investigation: AttentionLight (\textbf{Model 1}) and CoLight (\textbf{Model 2}).
Our experiments are structured around two configurations: 
\textbf{Config1}: uses queue length and current phase as the state, with negative queue length as the reward function; and \textbf{Config2}: uses the same state representations but with negative absolute pressure as the reward function.

\begin{figure}[htb]
    \centering
    \includegraphics[width=0.86\linewidth]{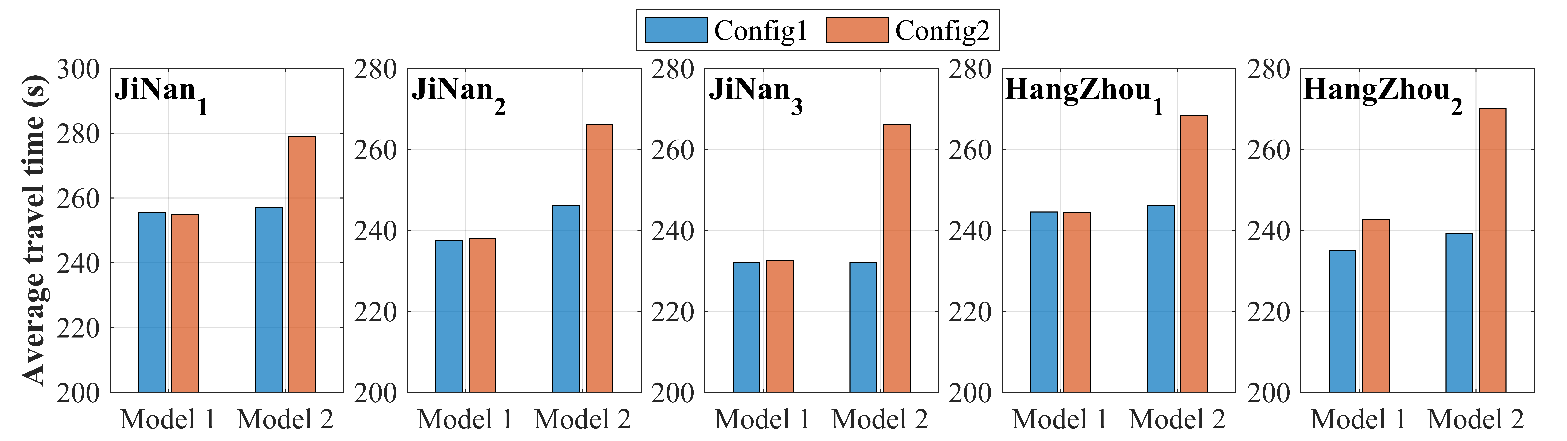}
    \caption{Model performance under different rewards w.r.t average travel time, the smaller the better.}
    \label{fig:reward}
\end{figure}

Experiments are conducted over JiNan and HangZhou, and the results are reported in Figure~\ref{fig:reward}.
Our findings show that AttentionLight performs slightly better under queue length than pressure.
Conversely, CoLight-based models  demonstrate significantly better performance under queue length compared to pressure.
%
In terms of state and reward calculation, queue length, which can be directly obtained from the traffic environment, is simpler to acquire than pressure, which necessitates complex calculation and neighbor information.
Therefore, using queue length as both the state and reward is a more favorable choice over pressure.
In conclusion, our experiments highlight the significance of selecting suitable reward functions in RL-based TSC approaches.
Utilizing queue length as both state and reward can enhance performance, especially in the case of CoLight-based models.

\subsection{Action Duration Study}

\begin{figure}[htb]
    \centering
    \includegraphics[width=0.86\linewidth]{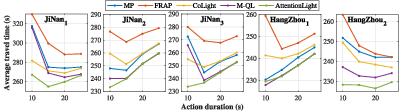}
    \caption{Model performance under different action duration.}
    \label{fig:duration}
\end{figure}

To evaluate the impact of action duration on model performance, we conduct extensive experiments using our proposed methods.
Figure~\ref{fig:duration} illustrates the model performance under different action duration. 
Our proposed AttentionLight consistently outperforms other methods over JiNan and HangZhou datasets.
Notably, M-QL exhibits better performance than FRAP and CoLight in most cases, suggesting that the traditional TSC methods remain powerful and essential.
The results highlight the crucial role action duration plays in the effectiveness of TSC models, and indicate that our proposed methods are robust and efficient under various action durations.

\subsection{Model Generalization}

Model generalization is a critical property of RL models, as an ideal RL model should be resilient to different traffic conditions after training in one traffic situation.
To evaluate the transferability of AttentionLight, We train it on JiNan and HangZhou datasets and transfer it to other datasets.
In each experiment, we calculated the average result of the final ten episodes.

\begin{figure}[htb]
    \centering
    \includegraphics[width=0.56\linewidth]{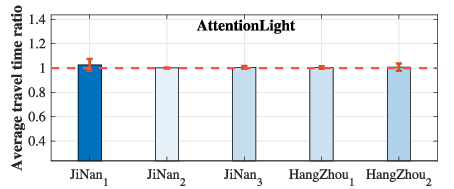}
    \caption{The average travel time of transfer divided by the average travel time of direct training. The error bars represent the 95\% confidence interval for the average travel time ratio.}
    \label{fig:transfer}
\end{figure}

%

%
The transfer performance is denoted as the average travel time ratio: $\frac{t_{transfer}}{t_{train}}$, where $t_{transfer}$ and  $t_{train} $ represent the average travel time of transfer and direct training, respectively.
The closer the average travel time ratio is to one, the less degradation is caused when facing a new environment.
Figure~\ref{fig:transfer} demonstrates the transferability of AttentionLight on JiNan and HangZhou, indicating that its model generalization is of great significance.
AttentionLight achieved high transfer performance over all datasets, suggesting that it is highly adaptable to new traffic environments.


\section{Conclusion}

In this paper, we propose the use of queue length as an efficient state representation for TSC and present two novel methods: Max Queue-Length (M-QL) and AttentionLight.
M-QL is an optimization-based method that is built on queue length, and AttentionLight uses self-attention to learn the phase correlation without requiring human knowledge.
Our proposed methods outperform previous state-of-the-art methods, with AttentionLight achieving the best performance. 
Furthermore, our experiments highlight the importance of state representation in addition to neural network design for RL.


%
However, we acknowledge that queue length alone may not be sufficient for complex traffic conditions,  and additional information should be incorporated into the state representation.
In future research, we aim to explore the inclusion of more information about the traffic conditions in the RL agent observations.
Additionally, we aim to investigate the use of more complex reward functions and network structures to further improve the performance of TSC.


\subsubsection{Acknowledgements}
This work was supported by grants from the National Natural Science Foundation of China (32225032, 32001192, 31322010, 32271597, 42201041), the Innovation Base Project of Gansu Province (20190323), the Top Leading Talents in Gansu Province to JMD, the National Scientific and Technological Program on Basic Resources Investigation (2019FY102002).

%
%
%
\bibliographystyle{splncs04}
\bibliography{mybib}

\end{document}